\documentclass[final,3p,times,twocolumn]{elsarticle}

\usepackage{epsfig}

\usepackage{graphicx}%
\usepackage{multirow}%
\usepackage{amsmath,amssymb,amsfonts}%
\usepackage{amsthm}%
\usepackage{mathrsfs}%
\usepackage[title]{appendix}%
\usepackage{textcomp}%
\usepackage{manyfoot}%
\usepackage{booktabs}%
\usepackage{algorithmicx}%
\usepackage{algpseudocode}%
\usepackage{listings}%
\usepackage[ruled,vlined]{algorithm2e}
\usepackage{color, colortbl}
\usepackage{siunitx}

\usepackage{booktabs}
\usepackage{multirow}
\usepackage[table,xcdraw]{xcolor}
\usepackage{times,epsf,mathtools,xcolor,amssymb,amsmath,hyperref, enumitem,wrapfig, subcaption, comment, cleveref, appendix, float}
\usepackage{tabu}
\usepackage{xcolor}
\usepackage{graphicx}
\usepackage{tikz}
\usetikzlibrary{shapes.geometric, arrows}

\definecolor{LightCyan}{rgb}{0.88,1,1}

\begin{document}

\begin{frontmatter}



\title{ABD-TDA:Attention-Based Deep and Topological Feature Fusion for Histopathological Cancer Classification}

\author[aff1]{Faisal Ahmed\corref{cor1}}
\ead{ahmedf9@erau.edu}

 \cortext[cor1]{Corresponding author}

 \address[aff1]{Department of Data Science and Mathematics, Embry-Riddle Aeronautical University, 3700 Willow Creek Rd, Prescott, Arizona 86301, USA}

\begin{abstract}
Accurate histopathological image classification remains challenging due to substantial morphological and structural variability across cancer types. While Vision Transformers (ViTs) capture global visual information through self-attention, they do not explicitly represent the topological organization of tissue structures. We propose \textbf{ABD-TDA}, a deep and topological feature fusion framework that combines transformer-based visual features with complementary topological descriptors. A pretrained ViT extracts a 768-dimensional visual representation, while a dedicated topological branch processes 1,200 Betti-$0$ and Betti-$1$ features derived from RGB and HSV color spaces. The resulting 64-dimensional topological representation is fused with the ViT features for classification.

We evaluate ABD-TDA on the ICIAR2018 breast cancer and UT-Osteosarcoma bone cancer datasets, containing four and three classes, respectively. The framework achieves 94.10\% and 96.54\% accuracy, with macro-AUC scores of 99.42\% and 99.65\%, respectively. Ablation results show consistent improvements from feature fusion, increasing accuracy from 92.67\% with ViT alone to 94.10\% on ICIAR2018 and from 94.15\% to 96.54\% on UT-Osteosarcoma. These results demonstrate that topological descriptors provide complementary structural information to transformer-based visual features, leading to more discriminative representations for histopathological cancer classification.
\end{abstract}


\begin{highlights}
    \item A novel deep-topological feature fusion framework, termed \textbf{ABD-TDA}, is proposed for histopathological cancer classification.

    \item The framework integrates Vision Transformer (ViT) features with \textbf{1,200 Betti-$0$ and Betti-$1$ descriptors} extracted from RGB and HSV representations.

    \item The proposed approach is evaluated on ICIAR2018 breast cancer and UT-Osteosarcoma bone cancer datasets, covering diverse histopathological tissue morphologies.

    \item ABD-TDA achieves \textbf{94.10\%} and \textbf{96.54\%} classification accuracy on the breast and bone cancer datasets, respectively, outperforming the compared state-of-the-art methods.

    \item Ablation experiments demonstrate that incorporating topological features consistently improves ViT performance, confirming the complementary value of deep visual and topological representations.
\end{highlights}


\end{frontmatter}




\section{Introduction}
\label{sec:introduction}

Cancer remains a major global health challenge, with a substantial and continually increasing burden on healthcare systems worldwide. Recent GLOBOCAN estimates reported more than 20 million new cancer cases and approximately 9.8 million cancer-related deaths globally in 2024, emphasizing the need for accurate and efficient diagnostic technologies~\cite{sung2026global}. Among different cancer types, breast and bone cancers pose considerable diagnostic challenges due to heterogeneous morphological patterns, complex tissue organization, and variations in cellular and structural characteristics. Accurate histopathological classification is therefore important for diagnosis, treatment planning, and disease management.

Histopathological examination remains a fundamental component of cancer diagnosis. In conventional practice, stained tissue specimens are digitized and examined microscopically by expert pathologists. However, manual analysis is time-consuming and requires substantial domain expertise, while subtle morphological differences can introduce interobserver variability. The increasing availability of whole-slide imaging (WSI) has therefore encouraged the development of computational pathology methods that can assist pathologists through automated and reproducible image analysis~\cite{komura2018machine,tschandl2020human}.

Deep learning (DL), particularly convolutional neural networks (CNNs), has significantly advanced automated histopathological image analysis. CNN-based approaches can learn hierarchical representations directly from tissue images and have been successfully applied to cancer detection, classification, grading, and prognosis prediction~\cite{litjens2017survey,esteva2019guide}. Nevertheless, conventional CNNs primarily focus on local spatial patterns, which can limit their ability to capture long-range dependencies and global tissue organization. This limitation is particularly relevant to histopathological images, where diagnostically meaningful information may occur across multiple spatial scales.

Vision Transformers (ViTs) have emerged as an effective alternative to convolution-based architectures for computational histopathology. Through self-attention, ViTs model relationships among image patches and capture global contextual information across the input image. Recent studies have demonstrated the effectiveness of transformer-based architectures for histopathological classification and other computational pathology tasks~\cite{dosovitskiy2020image,xu2023vision}. More recently, pathology foundation models such as UNI, CONCH, and Virchow have further highlighted the potential of transformer-based representations for learning transferable features across tissues and diagnostic tasks~\cite{chen2024towards,lu2024visual,vorontsov2024foundation}. Despite these advances, image-based deep representations do not explicitly encode the structural and topological properties of tissue organization. Related studies have also explored transfer learning, transformer architectures, and alternative visual representations for medical and histopathological image analysis~\cite{ahmed2025colormap,ahmed2026hog,ahmed2025ocuvit,ahmed2025histovit,ahmed2025repvit,ahmed2026pseudocolorvit,rawat2025efficient,ahmed2025transfer,ahmed2026bridging,ahmed2026enhancing}.

Topological data analysis (TDA) offers a complementary framework for characterizing the structural organization of complex data. Unlike conventional appearance-based descriptors, topological representations capture properties related to connectivity, components, and higher-order structures that can remain informative under certain geometric variations. TDA has increasingly been investigated in medical image analysis, including retinal imaging, brain imaging, and histopathological analysis~\cite{ahmed2026four,ahmed2025topo,ahmed2023topo,ahmed2023topological,ahmed2023tofi,ahmed20253d,yadav2023histopathological,ahmed2025robust,ahmed2026topological,ahmed2026brain,ahmed2026hybrid}. For histopathological images, such information can complement deep visual features by providing explicit descriptors of tissue organization. In particular, Betti-$0$ characterizes connected components, whereas Betti-$1$ captures one-dimensional topological structures such as loops or holes. These properties make TDA a promising complementary representation for histopathological classification.

Motivated by the complementary characteristics of deep visual and topological representations, we propose \textbf{ABD-TDA}, a deep-topological feature fusion framework for histopathological cancer classification. The framework combines high-level visual features extracted using a Vision Transformer with explicit topological descriptors derived from histopathological images. Specifically, we extract 1,200 Betti-$0$ and Betti-$1$ features from RGB and HSV representations. These features are processed through a dedicated multilayer perceptron (MLP) branch and fused with the 768-dimensional ViT representation. The resulting joint representation enables the classifier to exploit complementary visual and structural information.

We evaluate ABD-TDA on two publicly available benchmarks representing distinct cancer types: ICIAR2018 for four-class breast cancer classification and UT-Osteosarcoma for three-class bone cancer classification. In comparison with existing methods, ABD-TDA achieves 94.10\% accuracy on ICIAR2018 and 96.54\% on UT-Osteosarcoma. The ablation study further shows that fusing ViT and TDA improves performance over the individual branches, achieving 94.10\% and 96.54\% accuracy on the breast and bone datasets, respectively. These results demonstrate that explicit topological information can provide complementary structural cues to transformer-based visual representations.

The main contributions of this study are summarized as follows:

\begin{itemize}

    \item We propose \textbf{ABD-TDA}, a deep-topological feature fusion framework that combines Vision Transformer features with explicit topological descriptors for histopathological cancer classification.

    \item We extract 1,200 Betti-$0$ and Betti-$1$ features from RGB and HSV representations to capture complementary structural and topological characteristics of histopathological images.

    \item We employ a dedicated TDA feature-processing branch and fuse its learned representation with the 768-dimensional ViT representation, enabling joint learning from visual and topological information.

    \item We evaluate the proposed framework on two histopathological benchmarks, ICIAR2018 breast cancer and UT-Osteosarcoma bone cancer, demonstrating its applicability across different tissue types and classification tasks.

    \item Ablation experiments demonstrate that integrating TDA with ViT improves precision, recall, F1-score, accuracy, and AUC compared with the individual ViT and TDA configurations.

    \item ABD-TDA achieves 94.10\% accuracy on ICIAR2018 and 96.54\% accuracy on UT-Osteosarcoma in comparison with existing methods.

\end{itemize}

\section{Related Work}
\label{sec:related}

\subsection{Deep Learning for Histopathological Image Analysis}

Deep learning (DL) has substantially advanced computational pathology, particularly for automated histopathological image analysis, cancer detection, grading, and classification~\cite{litjens2017survey,esteva2019guide}. Convolutional neural networks (CNNs) have been extensively adopted because of their ability to automatically learn hierarchical representations from histological images, ranging from low-level cellular patterns to higher-level tissue characteristics~\cite{shen2017deep}. CNN-based approaches have demonstrated strong performance in tumor localization, tissue segmentation, cancer classification, and other pathology-related tasks. However, conventional CNNs primarily rely on local convolutional operations, which can make the modeling of long-range spatial dependencies and global tissue organization challenging. In addition, computational pathology systems often require large annotated datasets and may be sensitive to staining variations, tissue heterogeneity, and differences in acquisition protocols~\cite{komura2018machine,tschandl2020human}.

The limitations of conventional CNN-based approaches have motivated the exploration of architectures capable of modeling broader contextual relationships within histopathological images. In particular, the large spatial extent and structural complexity of whole-slide and patch-based pathology images make global representation learning an important consideration. Consequently, recent research has increasingly investigated attention-based and transformer-based architectures as alternatives or complements to convolutional models.

\subsection{Vision Transformers in Computational Pathology}

The introduction of the Vision Transformer (ViT) demonstrated that transformer architectures can effectively learn visual representations by treating an image as a sequence of patches and modeling their relationships through self-attention~\cite{dosovitskiy2020image}. Unlike conventional CNNs, which progressively construct receptive fields through local convolutions, ViTs can directly model relationships between distant image regions. Data-efficient transformer variants such as DeiT further demonstrated that transformer-based visual models can be trained effectively under more limited data regimes~\cite{touvron2021training}.

Transformers have subsequently been adopted across medical image analysis for classification, segmentation, and multimodal learning~\cite{chen2021transunet,rao2021global,wang2022transformer}. Their ability to capture long-range dependencies is particularly relevant to histopathology, where diagnostic information can be distributed across cellular, glandular, and tissue-level structures. Recent studies and surveys have specifically highlighted the growing role of vision transformers in computational histopathology~\cite{attallah2023vision,zhao2023recent,zhou2024comprehensive, xu2023vision}.

The development of pathology-specific foundation models has further demonstrated the potential of transformer-based representations. For example, UNI introduced a general-purpose pathology foundation model capable of learning transferable representations from large-scale histopathology data~\cite{chen2024towards}. CONCH extended this direction through vision-language pretraining for computational pathology~\cite{lu2024visual}, while Virchow demonstrated the effectiveness of large-scale self-supervised learning for general-purpose pathology representation learning~\cite{vorontsov2024foundation}. These developments suggest that transformer-based architectures can provide rich semantic representations across heterogeneous tissue types and diagnostic tasks.

Several recent studies have also explored transfer learning, transformer architectures, and alternative visual representations for medical and histopathological image analysis~\cite{ahmed2025colormap,ahmed2026hog,ahmed2025ocuvit,ahmed2025histovit,ahmed2025repvit,ahmed2026pseudocolorvit,rawat2025efficient,ahmed2025transfer,ahmed2026bridging,ahmed2026enhancing}. Collectively, these studies demonstrate the increasing effectiveness of transformer-based representation learning for medical image classification. Nevertheless, most transformer-based approaches primarily learn appearance and semantic representations directly from image data and do not explicitly incorporate topological information describing tissue organization.

\subsection{Topological Data Analysis in Medical and Histopathological Imaging}

Topological Data Analysis (TDA) provides a complementary mathematical framework for characterizing structural properties of complex data. Unlike conventional intensity-, texture-, or deep feature-based representations, TDA focuses on properties such as connectivity, components, loops, and other structural relationships. Persistent homology is particularly useful for extracting robust topological descriptors from high-dimensional data, with Betti numbers providing interpretable summaries of topological structures.

TDA has increasingly been explored in medical image analysis, including applications in retinal imaging, brain imaging, and histopathological analysis~\cite{ahmed2026four,ahmed2025topo,ahmed2023topo,ahmed2023topological,ahmed2023tofi,ahmed20253d,yadav2023histopathological,ahmed2025robust,ahmed2026topological,ahmed2026brain,ahmed2026hybrid}. These studies demonstrate that topological descriptors can capture structural characteristics that may not be adequately represented by conventional image features alone. In particular, topological information can provide complementary descriptions of tissue organization and morphological structure, making it potentially valuable for histopathological classification.

The usefulness of TDA is especially relevant to histopathology because tissue morphology is inherently structural. Cellular organization, connected tissue regions, glandular structures, and spatial relationships can contain important diagnostic information. However, TDA-based representations alone may not capture the rich semantic and appearance-based information learned by modern deep neural networks. This motivates the integration of topological descriptors with learned visual representations rather than treating the two approaches as competing alternatives.

\subsection{Deep and Topological Feature Fusion}

Recent advances in computational pathology suggest that combining complementary representations can improve the robustness of automated diagnosis. Deep learning models can learn highly discriminative semantic representations directly from image data, whereas hand-crafted or mathematically derived descriptors can provide additional structural information. Transformer-based pathology models further demonstrate the benefits of learning representations that capture both local morphology and broader tissue context~\cite{chen2024towards,lu2024visual,vorontsov2024foundation}.

Despite these advances, relatively limited attention has been given to explicitly combining transformer-based visual representations with topological descriptors for histopathological cancer classification. In particular, the integration of Betti-based features extracted from multiple color representations with transformer features remains insufficiently explored. This creates an opportunity to investigate whether explicit topological information can complement the global visual representations learned by a Vision Transformer.

\subsection{Positioning of the Proposed Method}

Motivated by the complementary characteristics of deep visual and topological representations, we propose \textbf{ABD-TDA}, a deep-topological feature fusion framework for histopathological cancer classification. Unlike approaches that rely exclusively on CNN or transformer representations, the proposed framework combines a fine-tuned Vision Transformer with an independent TDA branch. Specifically, 1,200 Betti-$0$ and Betti-$1$ descriptors are extracted from RGB and HSV representations and transformed through a dedicated multilayer perceptron before being fused with the 768-dimensional ViT representation.

The proposed framework is evaluated on two histopathological cancer datasets, ICIAR2018 for breast cancer and UT-Osteosarcoma for bone cancer. In addition to comparison with existing methods, an ablation study is performed to quantify the individual and combined contributions of ViT and TDA representations. The experimental results demonstrate that the fusion of deep visual and topological features consistently improves classification performance over the individual branches, supporting the hypothesis that explicit structural information provides complementary information to transformer-based visual representations.

\begin{figure*}[t!]
    \centering
    \includegraphics[width=\linewidth]{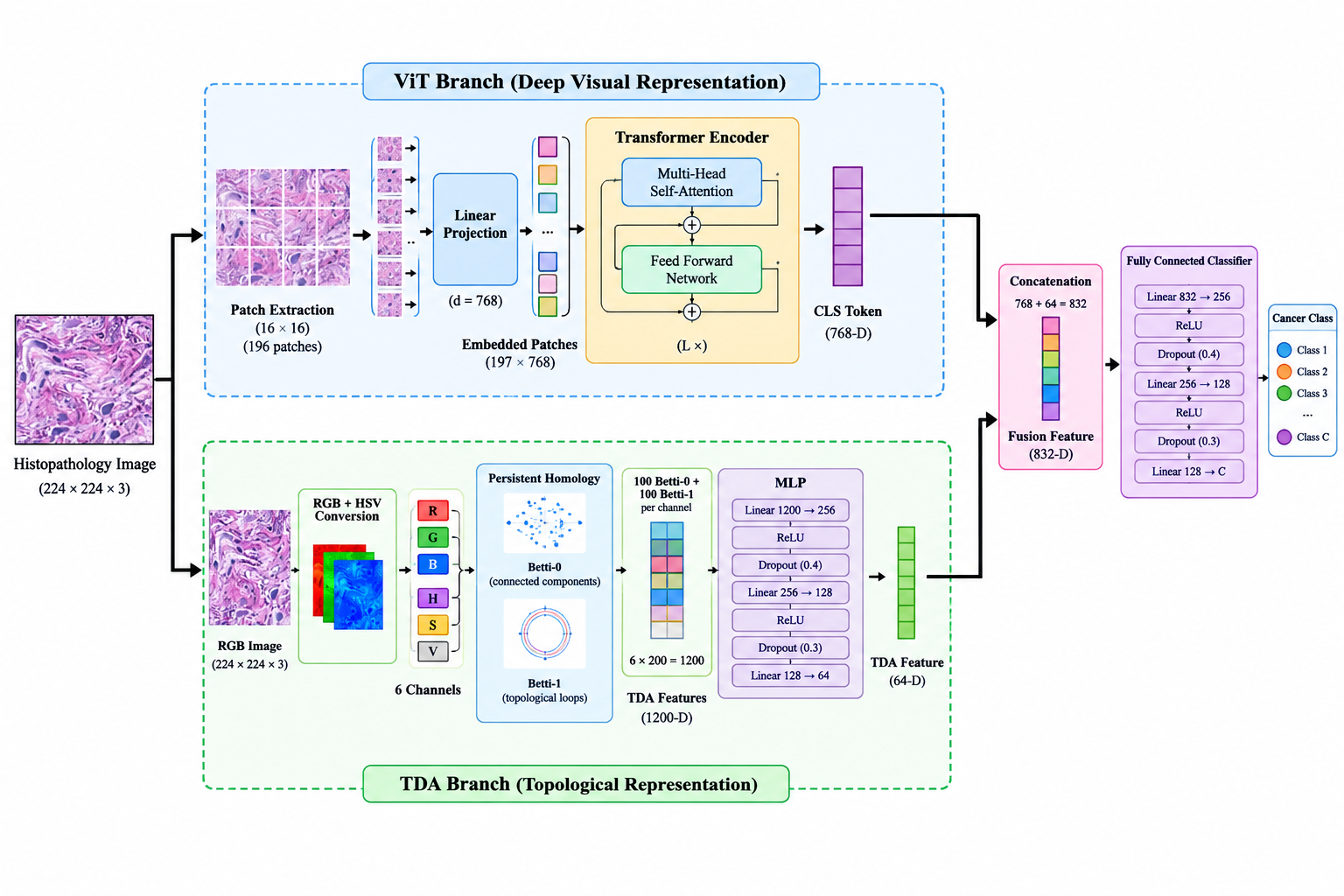}
    \caption{\footnotesize
    \textbf{Flowchart of the proposed ABD-TDA framework.}
    The framework consists of two complementary branches. The ViT branch extracts deep visual features from image patches using a pretrained Vision Transformer and obtains a 768-dimensional CLS representation. In parallel, the TDA branch converts the image into six RGB and HSV channels and extracts Betti-$0$ and Betti-$1$ features, producing a 1200-dimensional topological representation. The TDA features are reduced to 64 dimensions using an MLP. The 768-dimensional ViT and 64-dimensional TDA features are then concatenated to form an 832-dimensional fused representation, which is passed through a fully connected classifier to predict the cancer class.
    }
    \label{fig:flowchart}
\end{figure*}

\section{Methodology}
\label{sec:method}

This section presents the proposed \textbf{ABD-TDA} framework for histopathological cancer classification. The framework combines transformer-based deep visual representations with complementary topological descriptors extracted from multiple color representations of histopathological images. The overall architecture consists of three main components: (i) Vision Transformer (ViT) feature extraction, (ii) multi-color-space Topological Data Analysis (TDA) feature extraction, and (iii) deep-topological feature fusion followed by classification. Unlike approaches relying exclusively on learned visual features, ABD-TDA explicitly incorporates structural information through Betti-$0$ and Betti-$1$ descriptors extracted independently from the RGB and HSV channels.

\subsection{Vision Transformer Feature Extraction}
\label{subsec:vit_features}

Let
\begin{equation}
\mathbf{I} \in \mathbb{R}^{H \times W \times 3}
\end{equation}
denote an RGB histopathological image. Each input image is first resized to a spatial resolution of $224\times224$ pixels to satisfy the input requirement of the pretrained ViT-Base model. The pixel values are then scaled to the interval $[0,1]$ as

\begin{equation}
\mathbf{I}_{\mathrm{norm}}
=
\frac{\mathbf{I}}{255},
\qquad
\mathbf{I}_{\mathrm{norm}}
\in
[0,1]^{224\times224\times3}.
\end{equation}

Unlike grayscale medical images that require channel replication, the histopathological images used in this study are naturally represented using three RGB channels and can therefore be provided directly to the ViT after resizing and normalization.

The proposed framework employs the pretrained
\texttt{google/vit-base-patch16-224} architecture. Each normalized image is divided into non-overlapping patches of size $16\times16$. Therefore, the number of image patches is

\begin{equation}
N
=
\frac{224\times224}{16\times16}
=
196.
\end{equation}

The input image can consequently be represented as a sequence of patches,

\begin{equation}
\mathbf{I}_{\mathrm{norm}}
\rightarrow
\left\{
\mathbf{x}_{1},
\mathbf{x}_{2},
\dots,
\mathbf{x}_{196}
\right\}.
\end{equation}

Each image patch is flattened and linearly projected into the transformer embedding space:

\begin{equation}
\mathbf{e}_{i}
=
\mathbf{W}_{e}
\operatorname{vec}(\mathbf{x}_{i})
+
\mathbf{b}_{e},
\end{equation}

where $\mathbf{W}_{e}$ and $\mathbf{b}_{e}$ represent the learnable projection parameters. A learnable classification token, denoted by $\mathbf{x}_{\mathrm{CLS}}$, is prepended to the sequence, and positional embeddings are added to preserve spatial information:

\begin{equation}
\mathbf{z}^{0}
=
[
\mathbf{x}_{\mathrm{CLS}},
\mathbf{e}_{1}+\mathbf{p}_{1},
\dots,
\mathbf{e}_{N}+\mathbf{p}_{N}
].
\end{equation}

The resulting sequence is propagated through the transformer encoder layers, which employ multi-head self-attention (MSA) and feed-forward networks (FFNs). The self-attention operation is expressed as

\begin{equation}
\operatorname{Attention}
(\mathbf{Q},\mathbf{K},\mathbf{V})
=
\operatorname{Softmax}
\left(
\frac{\mathbf{Q}\mathbf{K}^{T}}
{\sqrt{d_{k}}}
\right)
\mathbf{V},
\end{equation}

where $\mathbf{Q}$, $\mathbf{K}$, and $\mathbf{V}$ denote the query, key, and value matrices, respectively, and $d_k$ represents the dimensionality of the key vectors.

Following the final transformer encoder layer, the representation associated with the classification token is extracted:

\begin{equation}
\mathbf{f}_{\mathrm{ViT}}
=
\mathbf{z}^{L}_{\mathrm{CLS}},
\end{equation}

where

\begin{equation}
\mathbf{f}_{\mathrm{ViT}}
\in
\mathbb{R}^{768}.
\end{equation}

Thus, each histopathological image is represented by a 768-dimensional deep feature vector encoding high-level semantic, morphological, and contextual information.

\begin{figure}[t!]
    \centering
    \includegraphics[width=\linewidth]{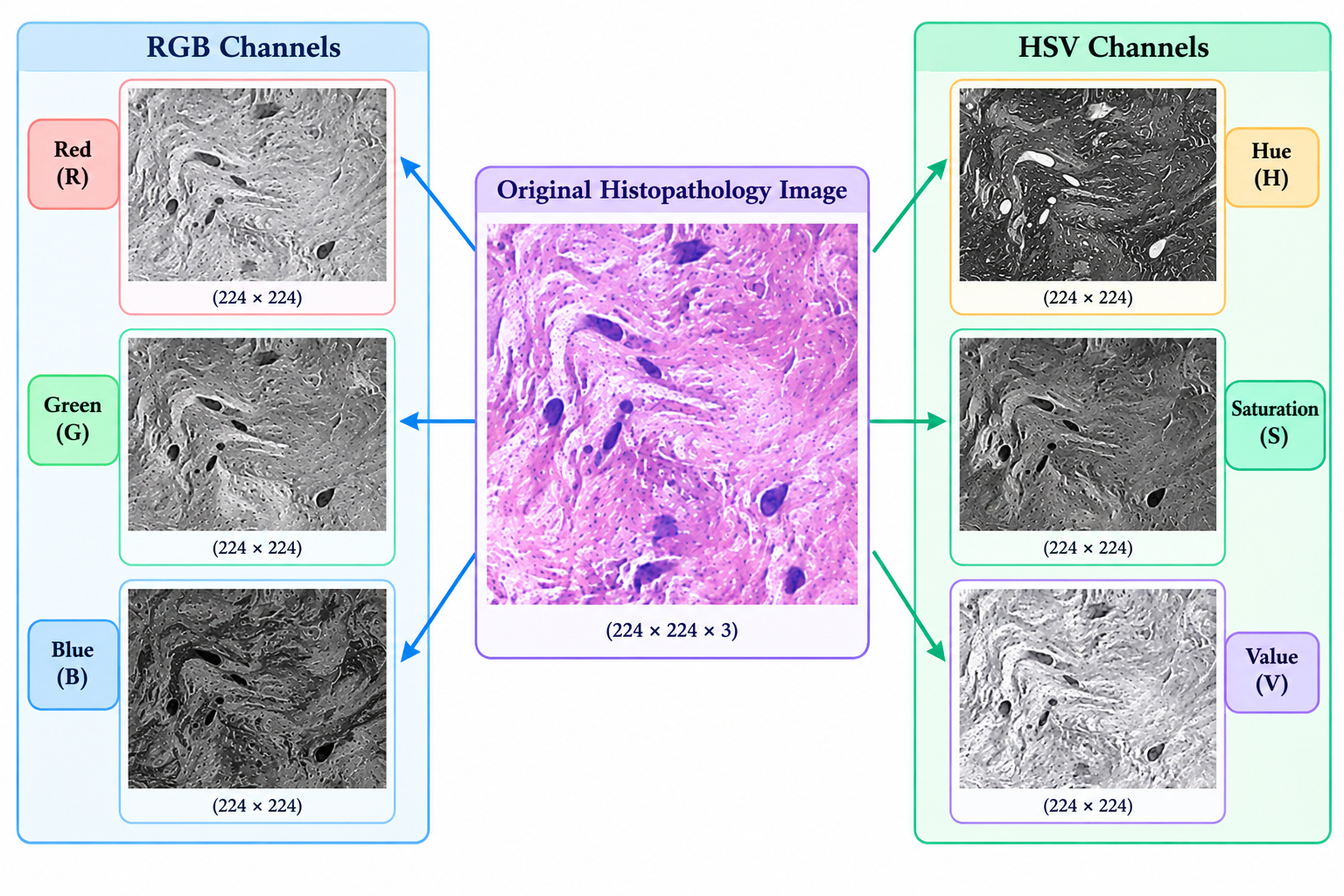}
    \caption{\footnotesize
    RGB and HSV Channel Extraction for Topological Feature Generation from Histopathological Images.
    }
    \label{fig:flowchart}
\end{figure}

\subsection{Multi-Color-Space Topological Feature Extraction}
\label{subsec:tda_features}

To complement the learned ViT representation, Topological Data Analysis is employed to explicitly characterize structural and connectivity patterns within the histopathological images. Rather than extracting topological information from a single grayscale representation, the proposed approach utilizes both \textbf{RGB} and \textbf{HSV} color spaces.

Given an RGB histopathological image, the six individual channel representations are defined as

\begin{equation}
\mathcal{C}
=
\{
R,G,B,H,S,V
\}.
\end{equation}

The RGB channels preserve the original color information associated with H\&E-stained tissue, whereas the HSV representation separates hue, saturation, and intensity-related characteristics. Processing both color representations enables the TDA branch to characterize complementary structural patterns appearing under different color decompositions.

For each channel $c\in\mathcal{C}$, let

\begin{equation}
\mathbf{X}^{(c)}
\in
\mathbb{R}^{H\times W}
\end{equation}

represent the corresponding two-dimensional channel image. A cubical filtration is constructed according to channel intensity values, producing a nested sequence of cubical complexes,

\begin{equation}
\mathcal{X}^{(c)}_{1}
\subseteq
\mathcal{X}^{(c)}_{2}
\subseteq
\dots
\subseteq
\mathcal{X}^{(c)}_{T}.
\end{equation}

For a filtration threshold $t$, the corresponding cubical complex can be expressed as

\begin{equation}
\mathcal{X}^{(c)}(t)
=
\left\{
\Delta_{ij}
\subseteq
\mathbf{X}^{(c)}
\; \middle| \;
\gamma_{ij}^{(c)}
\leq t
\right\},
\end{equation}

where $\gamma_{ij}^{(c)}$ denotes the intensity associated with the cubical cell $\Delta_{ij}$ in channel $c$.

Persistent homology is subsequently computed across the filtration. For homological dimension $k$, the persistence diagram is represented as

\begin{equation}
PD_{k}^{(c)}
=
\left\{
(b_{i},d_{i})
\right\}_{i=1}^{m},
\end{equation}

where $b_i$ and $d_i$ denote the birth and death filtration values of the $i$th topological structure.

In this study, two homology dimensions are considered:

\begin{equation}
H_{0}
\quad \text{and} \quad
H_{1}.
\end{equation}

Here, $H_{0}$ characterizes connected components, whereas $H_{1}$ represents one-dimensional loops or holes. The corresponding Betti curve for homology dimension $k$ is defined as

\begin{equation}
\beta_{k}^{(c)}(t)
=
\sum_{i=1}^{m}
\mathbb{I}
\left(
b_{i}
\leq t
<
d_{i}
\right),
\end{equation}

where $\mathbb{I}(\cdot)$ denotes the indicator function.

For each image channel, the Betti-$0$ and Betti-$1$ representations are discretized into \textbf{100 features each}. Hence, the topological feature representation for a single channel is

\begin{equation}
\mathbf{f}^{(c)}_{\mathrm{TDA}}
=
[
\boldsymbol{\beta}^{(c)}_{0};
\boldsymbol{\beta}^{(c)}_{1}
]
\in
\mathbb{R}^{200},
\end{equation}

where

\begin{equation}
\boldsymbol{\beta}^{(c)}_{0}
\in
\mathbb{R}^{100},
\qquad
\boldsymbol{\beta}^{(c)}_{1}
\in
\mathbb{R}^{100}.
\end{equation}

Since six channels are analyzed independently, namely $R$, $G$, $B$, $H$, $S$, and $V$, the complete topological representation is obtained by concatenating the features from all channels:

\begin{equation}
\mathbf{f}_{\mathrm{TDA}}
=
[
\mathbf{f}^{(R)}_{\mathrm{TDA}};
\mathbf{f}^{(G)}_{\mathrm{TDA}};
\mathbf{f}^{(B)}_{\mathrm{TDA}};
\mathbf{f}^{(H)}_{\mathrm{TDA}};
\mathbf{f}^{(S)}_{\mathrm{TDA}};
\mathbf{f}^{(V)}_{\mathrm{TDA}}
].
\end{equation}

Therefore,

\begin{equation}
\mathbf{f}_{\mathrm{TDA}}
\in
\mathbb{R}^{1200},
\end{equation}

since

\begin{equation}
6
\times
(100_{\beta_{0}}+100_{\beta_{1}})
=
1200.
\end{equation}

This multi-color-space topological representation is designed to capture connectivity and structural variations that may not be explicitly encoded by the transformer-based visual representation.

\begin{figure*}[t!]
    \centering
    \includegraphics[width=\linewidth]{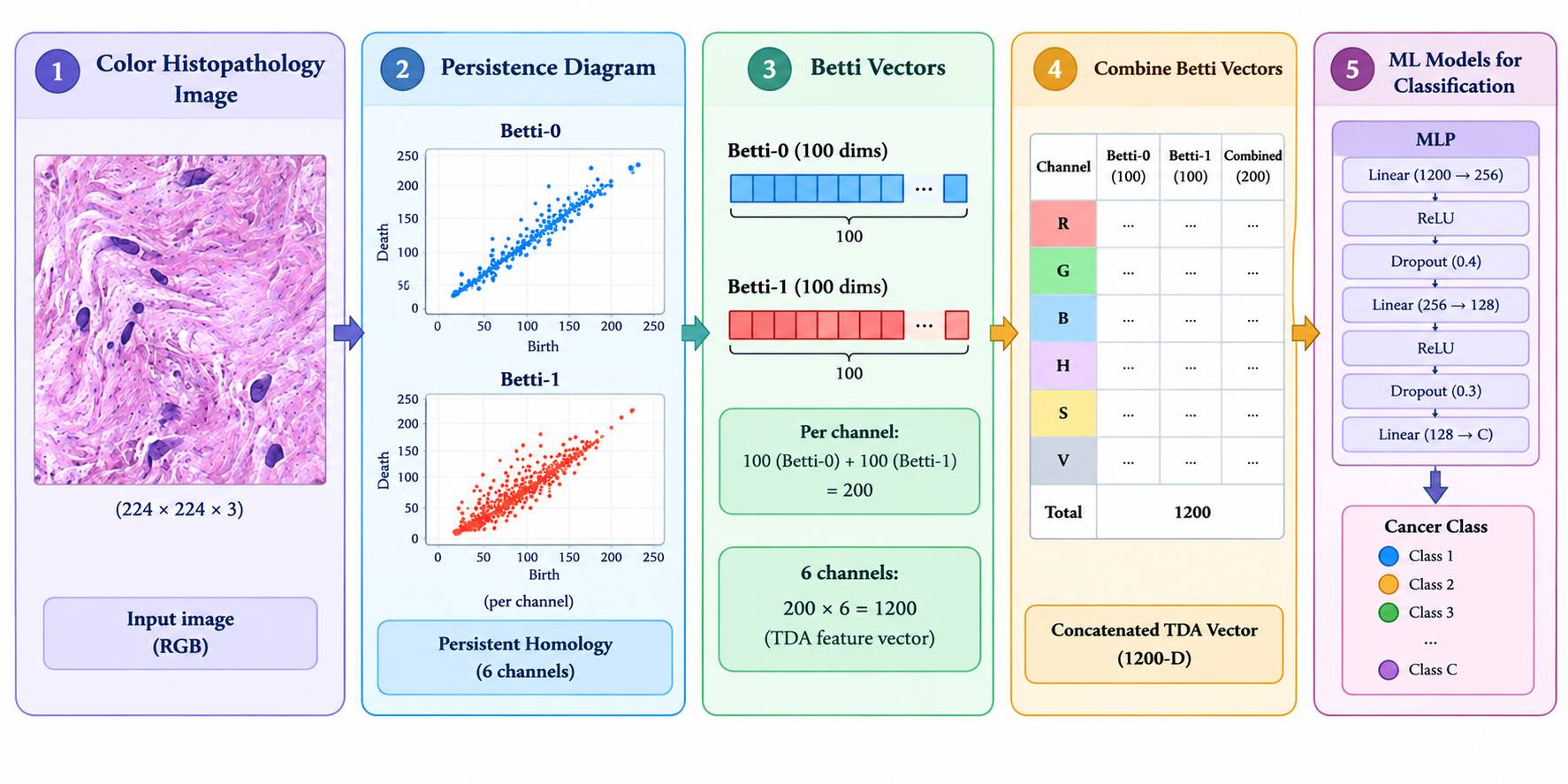}
    \caption{\footnotesize
    \textbf{TDA Feature Extraction Flowchart}
    }
    \label{fig:flowchart_tda}
\end{figure*}

\subsection{TDA Feature Learning Branch}
\label{subsec:tda_network}

Before fusion with the ViT representation, the 1,200-dimensional topological feature vector is standardized using statistics computed from the training data. The standardized TDA representation is then processed through a dedicated multilayer perceptron to obtain a compact task-specific embedding.

The first hidden layer is defined as

\begin{equation}
\mathbf{h}_{1}
=
\operatorname{ReLU}
\left(
\mathbf{W}_{1}
\mathbf{f}_{\mathrm{TDA}}
+
\mathbf{b}_{1}
\right),
\qquad
\mathbf{h}_{1}
\in
\mathbb{R}^{256}.
\end{equation}

A dropout probability of 0.3 is applied before the representation is propagated to the second layer:

\begin{equation}
\mathbf{h}_{2}
=
\operatorname{ReLU}
\left(
\mathbf{W}_{2}
\mathbf{h}_{1}
+
\mathbf{b}_{2}
\right),
\qquad
\mathbf{h}_{2}
\in
\mathbb{R}^{128}.
\end{equation}

A second dropout operation with probability 0.3 is followed by the final TDA embedding layer,

\begin{equation}
\mathbf{h}_{\mathrm{TDA}}
=
\operatorname{ReLU}
\left(
\mathbf{W}_{3}
\mathbf{h}_{2}
+
\mathbf{b}_{3}
\right),
\end{equation}

yielding

\begin{equation}
\mathbf{h}_{\mathrm{TDA}}
\in
\mathbb{R}^{64}.
\end{equation}

Thus, the dedicated TDA branch transforms the original 1,200-dimensional handcrafted topological descriptor into a compact 64-dimensional representation suitable for integration with the deep transformer features.

\subsection{Deep--Topological Feature Fusion}
\label{subsec:fusion}

The principal component of ABD-TDA is the integration of the deep representation learned by ViT with the compact topological representation learned by the TDA branch. Let

\begin{equation}
\mathbf{f}_{\mathrm{ViT}}
\in
\mathbb{R}^{768}
\end{equation}

denote the transformer representation and

\begin{equation}
\mathbf{h}_{\mathrm{TDA}}
\in
\mathbb{R}^{64}
\end{equation}

denote the learned topological representation. The two feature vectors are fused through feature-level concatenation:

\begin{equation}
\mathbf{f}_{\mathrm{fusion}}
=
[
\mathbf{f}_{\mathrm{ViT}};
\mathbf{h}_{\mathrm{TDA}}
].
\end{equation}

Consequently,

\begin{equation}
\mathbf{f}_{\mathrm{fusion}}
\in
\mathbb{R}^{832},
\end{equation}

where

\begin{equation}
832 = 768 + 64.
\end{equation}

The fused representation combines the global semantic and contextual information captured by the ViT with explicit structural and connectivity information encoded by persistent homology.

\subsection{Classification}
\label{subsec:classification}

The 832-dimensional fused representation is passed through a multilayer classification head. The first classification layer transforms the fused features into a 256-dimensional hidden representation:

\begin{equation}
\mathbf{z}_{1}
=
\operatorname{ReLU}
\left(
\mathbf{W}_{4}
\mathbf{f}_{\mathrm{fusion}}
+
\mathbf{b}_{4}
\right),
\qquad
\mathbf{z}_{1}
\in
\mathbb{R}^{256}.
\end{equation}

A dropout probability of 0.4 is applied, followed by a second fully connected layer:

\begin{equation}
\mathbf{z}_{2}
=
\operatorname{ReLU}
\left(
\mathbf{W}_{5}
\mathbf{z}_{1}
+
\mathbf{b}_{5}
\right),
\qquad
\mathbf{z}_{2}
\in
\mathbb{R}^{128}.
\end{equation}

A dropout probability of 0.3 is then applied before the final classification layer:

\begin{equation}
\mathbf{o}
=
\mathbf{W}_{6}
\mathbf{z}_{2}
+
\mathbf{b}_{6},
\end{equation}

where

\begin{equation}
\mathbf{o}
\in
\mathbb{R}^{C},
\end{equation}

and $C$ represents the number of diagnostic classes. In the present study,

\begin{equation}
C=
\begin{cases}
4, & \text{for ICIAR2018 (Breast Cancer)},\\
3, & \text{for UT-Osteosarcoma (Bone Cancer)}.
\end{cases}
\end{equation}

During evaluation, the output logits are converted into posterior class probabilities using the Softmax function:

\begin{equation}
\hat{y}_{c}
=
\frac{\exp(o_{c})}
{\displaystyle\sum_{j=1}^{C}\exp(o_{j})}.
\end{equation}

The predicted class is subsequently determined as

\begin{equation}
\hat{c}
=
\arg\max_{c}
\hat{y}_{c}.
\end{equation}

The complete network is optimized using the categorical cross-entropy loss:

\begin{equation}
\mathcal{L}
=
-\frac{1}{B}
\sum_{i=1}^{B}
\sum_{c=1}^{C}
y_{i,c}
\log
\left(
\hat{y}_{i,c}
\right),
\end{equation}

where $B$ denotes the batch size, $C$ represents the number of classes, $y_{i,c}$ is the ground-truth class indicator, and $\hat{y}_{i,c}$ denotes the corresponding predicted probability.

The proposed fusion mechanism therefore combines two fundamentally different representations of histopathological morphology. The ViT branch captures high-level semantic information and long-range visual dependencies through self-attention, while the TDA branch explicitly characterizes the evolution of connected components and loop structures across RGB and HSV image channels. Their integration provides an enriched feature representation that captures both learned visual characteristics and explicit topological organization. The effectiveness of this complementary fusion is further investigated through the ablation study in \Cref{tab:ablation}, where the standalone TDA, standalone ViT, and combined ViT+TDA configurations are compared.

\begin{table*}[htbp]
\centering
\caption{Class-wise distribution of training and test samples for the UT-Osteo and BACH histopathological cancer datasets.}
\label{tab:dataset_distribution}
\resizebox{0.85\textwidth}{!}{
\begin{tabular}{llrr}
\toprule
\textbf{Dataset} & \textbf{Class} & \textbf{Training Samples} & \textbf{Test Samples} \\
\midrule

\multirow{3}{*}{UT-Osteo (Bone)}
& Non-tumor         & 3,874 & 1,679 \\
& Non-viable tumor  & 2,696 & 1,131 \\
& Viable            & 3,447 & 1,495 \\
\cmidrule(lr){2-4}
& \textbf{Total}    & \textbf{10,017} & \textbf{4,305} \\

\midrule

\multirow{4}{*}{BACH (Breast)}
& Normal            & 3,248 & 1,402 \\
& Benign            & 3,324 & 1,428 \\
& In Situ           & 3,278 & 1,402 \\
& Invasive          & 3,339 & 1,427 \\
\cmidrule(lr){2-4}
& \textbf{Total}    & \textbf{13,189} & \textbf{5,659} \\

\bottomrule
\end{tabular}
}
\end{table*}

\section{Experiments}
\label{sec:experiments}

\subsection{Datasets}
\label{sec:datasets}

To evaluate the proposed ABD-TDA framework, we use two publicly available and widely adopted histopathological image benchmarks: the \textbf{BACH (ICIAR2018) Breast Cancer} dataset and the \textbf{UT-Osteosarcoma (UT-Osteo) Bone Cancer} dataset. To ensure a fair and reproducible comparison with previously published methods, we use the same training and test partitions adopted in the corresponding benchmark studies. The class-wise distributions of the training and test sets are presented in \Cref{tab:dataset_distribution}.

\textbf{BACH (Breast Cancer):}
The Breast Cancer Histology (BACH) dataset~\cite{aresta2019bach} was introduced as part of the ICIAR 2018 Grand Challenge on Breast Cancer Histology Images. It contains hematoxylin and eosin (H\&E)-stained breast histopathology images categorized into four classes: \textit{Normal, Benign, In Situ}, and \textit{Invasive}. The original BACH dataset contains 400 microscopy images, with 100 images per class. For our experiments, we use the corresponding image-tile dataset and retain the training and test partition used in previous studies to enable a consistent comparison with existing methods. As shown in \Cref{tab:dataset_distribution}, the training set contains 13,189 image tiles, while the test set contains 5,659 tiles, with a relatively balanced distribution across the four diagnostic categories. No data augmentation is employed in our experiments, allowing the evaluation to focus on the effectiveness of the proposed deep-topological feature representation.

\textbf{UT-Osteosarcoma (Bone Cancer):}
The UT-Osteosarcoma dataset~\cite{arunachalam2019viable} consists of H\&E-stained histopathological images of osteosarcoma collected by clinical researchers at the University of Texas Southwestern Medical Center. The dataset was constructed from archival samples obtained from 50 patients treated at Children's Medical Center, Dallas, between 1995 and 2015. The images are categorized into three classes based on their predominant tissue characteristics: \textit{Non-Tumor, Non-Viable Tumor}, and \textit{Viable Tumor}. Following the training and test partition used in previous studies, the dataset contains 10,017 training images and 4,305 test images. The training set comprises 3,874 Non-Tumor, 2,696 Non-Viable Tumor, and 3,447 Viable Tumor images, whereas the test set contains 1,679, 1,131, and 1,495 images for the respective classes. The class distribution is sufficiently balanced to support reliable multi-class evaluation without requiring additional class-level resampling.

Importantly, for both datasets, we retain the established training and test partitions rather than creating a new random split. This experimental setting ensures that the reported results are directly comparable with previously published approaches evaluated on the same benchmark partitions. It also reduces potential differences arising from alternative data-splitting strategies and provides a more consistent basis for assessing the effectiveness of the proposed ABD-TDA framework.

\subsection{Experimental Setup}
\label{sec:experimental_setup}

\noindent \textbf{Training--Test Protocol:}
To ensure a fair and consistent comparison with previously published methods, we use the same established training--test partitions adopted in the corresponding studies reported in our comparative tables. We do not introduce a new random partition of the datasets. For the \textbf{UT-Osteo (bone cancer)} dataset, we follow the commonly reported 70:30 training--test protocol, while for the \textbf{ICIAR2018 (BACH) breast cancer} dataset, we follow the established 80:20 training--test protocol. The exact class-wise numbers of training and test samples used in our experiments are reported in \Cref{tab:dataset_distribution}. This protocol ensures that the performance of ABD-TDA is evaluated under the same data-partition setting as the compared methods, thereby providing a more reliable and reproducible basis for performance comparison. Both datasets are formulated as multi-class classification problems, with three classes for UT-Osteo and four classes for ICIAR2018.
 
\smallskip

\noindent \textbf{No Data Augmentation:}
No data augmentation techniques are applied during training. The model is trained exclusively on the original training images, without introducing additional samples through geometric, color, or intensity transformations. This setting provides a consistent evaluation of the proposed ABD-TDA framework and allows its performance to be attributed to the learned deep and topological representations rather than to augmented training data.

\smallskip

\noindent \textbf{Model Hyperparameters:}
The proposed ABD-TDA framework is implemented using the pre-trained \texttt{google/vit-base-patch16-224} Vision Transformer from the Hugging Face Transformers library. The ViT produces a 768-dimensional representation from each $224\times224$ input image. The 1,200-dimensional TDA representation is processed through an MLP with three fully connected layers, reducing the feature dimension from 1,200 to 64. The resulting 64-dimensional topological representation is concatenated with the 768-dimensional ViT representation, producing an 832-dimensional fused feature vector. This representation is subsequently processed by the classification head. The complete training configuration is summarized in \Cref{tab:hyperparameters}. The model is optimized using AdamW with a learning rate of $1\times10^{-4}$ and weight decay of $1\times10^{-4}$, using cross-entropy loss and a batch size of 16. Training is performed for a maximum of 100 epochs with early stopping using a patience of 25 epochs.

\begin{table}[htbp]
\centering
\caption{Hyperparameter and training configuration of the proposed ABD-TDA framework.}
\label{tab:hyperparameters}
\resizebox{0.95\linewidth}{!}{
\begin{tabular}{ll}
\toprule
\textbf{Parameter} & \textbf{Configuration} \\
\midrule
Backbone & ViT-Base/16 \\
Pre-trained model & \texttt{google/vit-base-patch16-224} \\
ViT patch size & $16 \times 16$ \\
Input image size & $224 \times 224$ \\
ViT feature dimension & 768 \\
TDA input features & 1,200 \\
TDA descriptors & Betti-$0$ and Betti-$1$ \\
Color representations & RGB and HSV \\
TDA branch & MLP \\
TDA MLP architecture & $1200 \rightarrow 256 \rightarrow 128 \rightarrow 64$ \\
TDA activation & ReLU \\
TDA dropout & 0.30 \\
Fusion strategy & Feature concatenation \\
Fused feature dimension & $768 + 64 = 832$ \\
Classifier architecture & $832 \rightarrow 256 \rightarrow 128 \rightarrow 2/3/4$ \\
Classifier activation & ReLU \\
Classifier dropout & 0.40, 0.30 \\
Loss function & Cross-Entropy Loss \\
Optimizer & AdamW \\
Learning rate & $1 \times 10^{-4}$ \\
Weight decay & $1 \times 10^{-4}$ \\
Batch size & 16 \\
Maximum epochs & 50 \\
Early stopping patience & 25 epochs \\
Image scaling & Pixel values divided by 255 \\
Data augmentation & None \\
Hardware & GPU when available \\
\bottomrule
\end{tabular}
}
\end{table}

\smallskip


\vspace{-.1in}
\begin{figure*}[t!]
    \centering
    \subfloat[\scriptsize Confusion matrix for Bone Cancer classification.]
    {\includegraphics[width=0.42\linewidth]{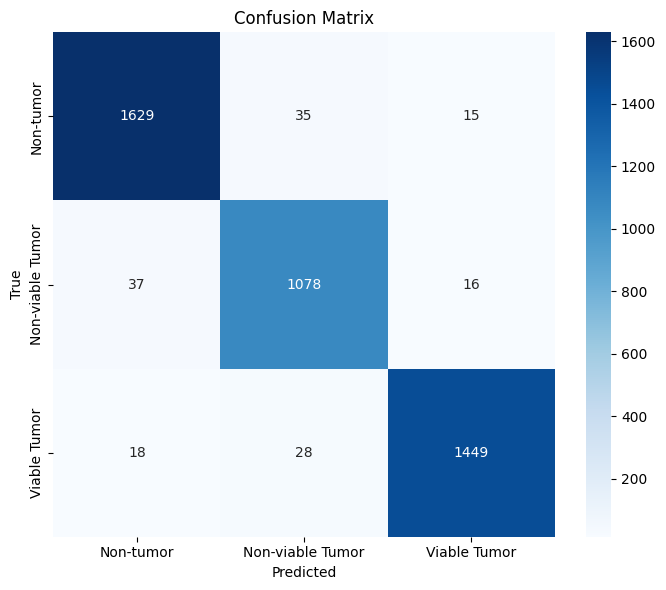}
    \label{fig:bone-matrix}}
    \hfill
    \subfloat[\scriptsize Confusion matrix for Breast Cancer classification.]
    {\includegraphics[width=0.42\linewidth]{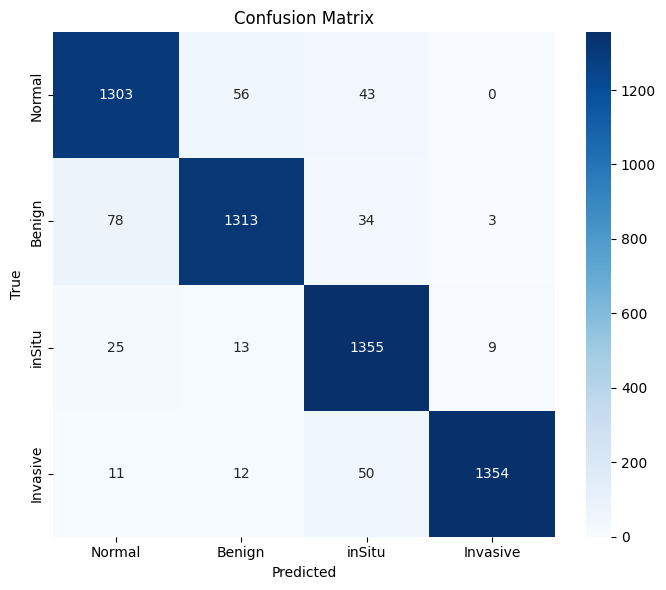}
    \label{fig:breast-matrix}}

    \caption{\footnotesize Confusion matrices obtained by the proposed ABD-TDA framework on the UT-Osteosarcoma (Bone Cancer) and ICIAR2018 (Breast Cancer) datasets. The matrices illustrate the distribution of true and predicted class labels, providing a detailed view of the model's classification performance across the diagnostic categories.}
    \label{fig:His-matrix}
\end{figure*}

\vspace{-.1in}
\begin{figure*}[t!]
    \centering
    \subfloat[\scriptsize ROC curve for Bone Cancer classification.]
    {\includegraphics[width=0.42\linewidth]{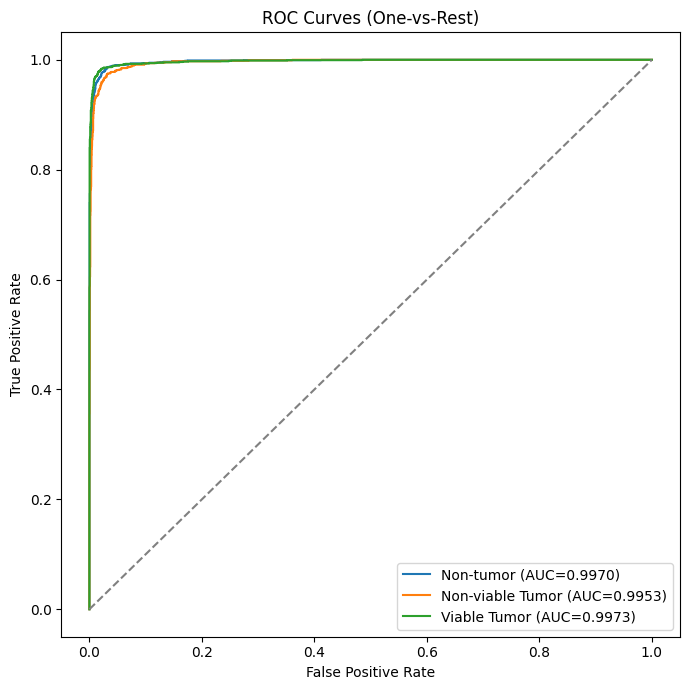}
    \label{fig:bone-roc}}
    \hfill
    \subfloat[\scriptsize ROC curve for Breast Cancer classification.]
    {\includegraphics[width=0.42\linewidth]{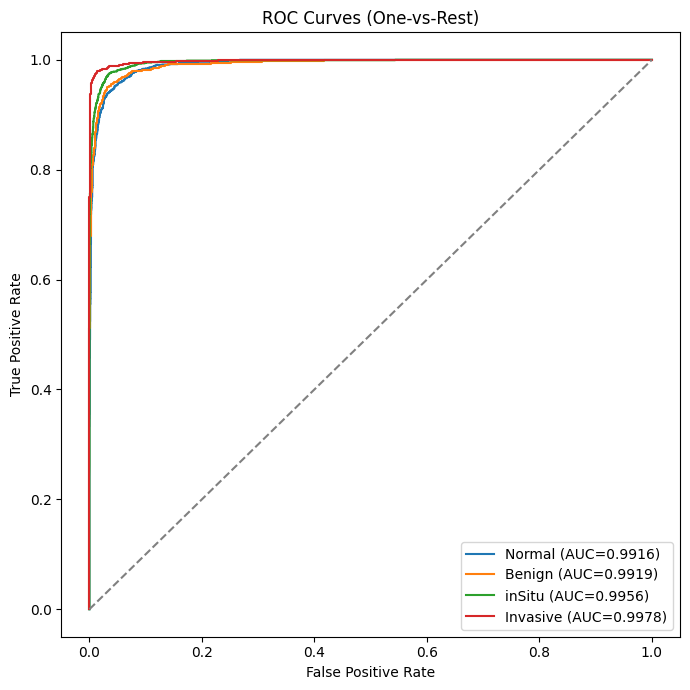}
    \label{fig:breast-roc}}

    \caption{\footnotesize Receiver operating characteristic (ROC) curves of the proposed ABD-TDA framework on the UT-Osteosarcoma (Bone Cancer) and ICIAR2018 (Breast Cancer) datasets. The ROC curves illustrate the discriminative performance of the proposed model across the respective diagnostic classes, with the area under the ROC curve (AUC) providing an overall measure of classification discrimination.}
    \label{fig:His-ROC}
\end{figure*}

\section{Results and Discussion}
\label{sec:results}

The proposed \textbf{ABD-TDA} framework was evaluated on the UT-Osteosarcoma and ICIAR2018 datasets using the same train--test settings as the corresponding comparative studies. The results are summarized in Tables~\ref{tab:bone} and~\ref{tab:breast}. For comparison, the TDA-LBP-Gabor method~\cite{yadav2023histopathological} uses 2,800 handcrafted features, comprising 1,600 TDA, 400 Gabor, and 800 LBP features. In contrast, our method extracts 1,200 topological features directly from the RGB and HSV channels and integrates them with ViT features.

\subsection{Bone Cancer Classification}
\label{subsec:bone_results}

Table~\ref{tab:bone} compares ABD-TDA with existing methods on UT-Osteosarcoma. Mishra-CNN~\cite{arunachalam2019viable}, Mishra-SVM~\cite{arunachalam2019viable}, VGG19~\cite{anisuzzaman2021deep}, and TDA-LBP-Gabor~\cite{yadav2023histopathological} achieve accuracies of 93.30\%, 89.90\%, 93.91\%, and 94.20\%, respectively. ABD-TDA achieves the highest accuracy of \textbf{96.54\%}, improving over TDA-LBP-Gabor by 2.34 percentage points and VGG19 by 2.63 percentage points.

The ablation results in Table~\ref{tab:ablation} further show the benefit of TDA fusion. ViT achieves 94.15\% accuracy, while ViT+TDA reaches \textbf{96.54\%}, with AUC increasing from 98.88\% to 99.65\%. These results indicate that the TDA features provide complementary structural information to the visual features learned by ViT.

\begin{table}[htbp]
\centering
\caption{Performance comparison of the proposed method with existing state-of-the-art methods on the UT-Osteosarcoma dataset for bone cancer classification.}
\label{tab:bone}
\resizebox{0.85\linewidth}{!}{%
\begin{tabular}{lccc}
\toprule
\textbf{Method} & \textbf{Train:Test} & \textbf{\# Classes} & \textbf{Accuracy (\%)} \\
\midrule
Mishra-CNN~\cite{arunachalam2019viable} & 70:30 & 3 & 93.30 \\
Mishra-SVM~\cite{arunachalam2019viable} & 70:30 & 3 & 89.90 \\
VGG19~\cite{anisuzzaman2021deep} & 70:30 & 3 & 93.91 \\
TDA-LBP-Gabor~\cite{yadav2023histopathological} & 70:30 & 3 & 94.20 \\
\midrule
\textbf{ABD-TDA (Ours)} & 70:30 & 3 & \textbf{96.54} \\
\bottomrule
\end{tabular}%
}
\end{table}

\subsection{Breast Cancer Classification}
\label{subsec:breast_results}

Table~\ref{tab:breast} presents the results on ICIAR2018. Previous methods report accuracies ranging from 79.00\% to 92.50\%, while the TDA-LBP-Gabor-based method~\cite{yadav2023histopathological} achieves 91.64\%. ABD-TDA obtains the highest accuracy of \textbf{94.10\%}, improving over DCNN~\cite{kassani2019breast} by 1.60 percentage points and TDA by 2.46 percentage points.

The ablation results show that ViT achieves 92.67\% accuracy, whereas ViT+TDA reaches \textbf{94.10\%}. The fused model also improves macro-precision from 92.72\% to 94.17\%, macro-recall from 92.64\% to 94.10\%, macro-F1 from 92.68\% to 94.13\%, and AUC from 99.14\% to 99.42\%. Overall, the results on both datasets demonstrate that combining ViT-based visual features with Betti-$0$ and Betti-$1$ topological features provides a more effective representation for histopathological cancer classification.

\begin{table}[htbp]
\centering
\caption{Performance comparison of the proposed method with existing state-of-the-art methods on the ICIAR2018 dataset for breast cancer classification.}
\label{tab:breast}
\resizebox{0.85\linewidth}{!}{%
\begin{tabular}{lccc}
\toprule
\textbf{Method} & \textbf{Train:Test} & \textbf{\# Classes} & \textbf{Accuracy (\%)} \\
\midrule
Kwok~\cite{kwok2018multiclass_breast5} & 75:25 & 4 & 79.00 \\
Nawaz~\cite{nawaz2018classification_breast3} & 80:20 & 4 & 81.25 \\
Rakhlin~\cite{guo2018_breast2} & 80:20 & 4 & 87.20 \\
Vang~\cite{vang2018deep_breast1} & 75:25 & 4 & 87.50 \\
DCNN~\cite{kassani2019breast} & 75:25 & 4 & 92.50 \\
TDA-LBP-Gabor~\cite{yadav2023histopathological} & 80:20 & 4 & 91.64 \\
\midrule
\textbf{ABD-TDA (Ours)} & 80:20 & 4 & \textbf{94.10} \\
\bottomrule
\end{tabular}%
}
\end{table}

\section{Ablation Study}
\label{sec:ablation}

An ablation study was performed to examine the contribution of the ViT and TDA components in the proposed ABD-TDA framework. Three configurations were evaluated: (i) MLP-TDA, which uses only the topological features, (ii) ViT, which uses only the deep visual representation, and (iii) ViT+TDA, which combines both representations. The experiments were conducted on the ICIAR2018 and UT-Osteosarcoma datasets. Macro-precision, macro-recall, accuracy, macro-F1, and macro-AUC were used for evaluation. The results are reported in Table~\ref{tab:ablation}.

\subsection{Contribution of Individual Components}
\label{subsec:ablation_components}

The standalone results show that ViT provides a stronger visual representation than the TDA-only configuration. On ICIAR2018, MLP-TDA achieves 83.50\% accuracy, compared with 92.67\% for ViT. Similarly, on UT-Osteosarcoma, the accuracy increases from 86.60\% with MLP-TDA to 94.15\% with ViT. This shows the importance of learned deep visual features for histopathological classification.

Although TDA alone produces lower classification accuracy, it still provides meaningful discriminative information. The MLP-TDA configuration achieves AUC values of 96.45\% and 96.26\% on the breast and bone datasets, respectively. Thus, the topological representation contains useful structural information even without the ViT branch.

\subsection{Effect of ViT--TDA Fusion}
\label{subsec:ablation_fusion}

The combination of ViT and TDA produces the strongest results on both datasets. For ICIAR2018, ViT+TDA improves accuracy from 92.67\% to 94.10\% and increases AUC from 99.14\% to 99.42\%. Macro-precision, macro-recall, and macro-F1 also increase after adding the TDA branch.

A larger improvement is observed for UT-Osteosarcoma. The fused model increases accuracy from 94.15\% with ViT alone to 96.54\%. The corresponding macro-precision, macro-recall, and macro-F1 scores reach 96.37\%, 96.42\%, and 96.39\%, respectively. The AUC also increases from 98.88\% to 99.65\%.

The consistent gains across the two datasets indicate that the TDA features provide complementary information rather than simply duplicating the information learned by ViT. The ViT branch learns high-level visual and contextual features, whereas the TDA branch describes structural properties through Betti-$0$ and Betti-$1$ features. Moreover, extracting these features from both RGB and HSV channels provides multiple representations of tissue structure.

Overall, the ablation study validates the central design of ABD-TDA. ViT provides the main discriminative visual representation, while TDA adds complementary topological information. Their fusion consistently improves classification performance across both histopathological datasets.

\begin{table*}[htbp]
\centering
\caption{Ablation study of the proposed deep and topological feature fusion framework on the ICIAR2018 and UT-Osteosarcoma datasets. All reported metrics are calculated using macro averaging. The best-performing configuration for each dataset is highlighted in bold.}
\label{tab:ablation}
\resizebox{\textwidth}{!}{%
\begin{tabular}{llcccccc}
\toprule
\textbf{Dataset} & \textbf{Model} & \textbf{\# Classes} &
\textbf{Precision (\%)} & \textbf{Recall (\%)} &
\textbf{F1-score (\%)} & \textbf{Accuracy (\%)} & \textbf{AUC (\%)} \\
\midrule

\multirow{3}{*}{ICIAR2018 (Breast)}
& MLP (TDA) & 4 & 83.55 & 83.46 & 83.50 & 83.50 & 96.45 \\
& ViT & 4 & 92.72 & 92.64 & 92.68 & 92.67 & 99.14 \\
& \textbf{ViT + TDA} & 4 & \textbf{94.17} & \textbf{94.10} &
\textbf{94.13} & \textbf{94.10} & \textbf{99.42} \\

\midrule

\multirow{3}{*}{UT-Osteosarcoma (Bone)}
& MLP (TDA) & 3 & 86.37 & 85.54 & 85.95 & 86.60 & 96.26 \\
& ViT & 3 & 93.98 & 94.17 & 94.07 & 94.15 & 98.88 \\
& \textbf{ViT + TDA} & 3 & \textbf{96.37} & \textbf{96.42} &
\textbf{96.39} & \textbf{96.54} & \textbf{99.65} \\

\bottomrule
\end{tabular}%
}
\end{table*}

\section{Limitations}
\label{sec:limitations}

Despite the promising performance of ABD-TDA, several limitations should be acknowledged. First, the evaluation is conducted on two benchmark datasets, which may not fully represent the diversity of real-world histopathological images. Second, the proposed framework introduces additional computational overhead due to the extraction and processing of topological features alongside ViT representations. Third, the current evaluation is limited to classification performance and does not assess external clinical validation or interpretability. Future work will investigate larger and more diverse multi-center datasets, computationally efficient TDA extraction, and clinically oriented validation and explainability methods.

\section{Conclusion}
\label{sec:conclusion}

In this study, we proposed \textbf{ABD-TDA}, a deep-topological feature fusion framework that integrates Vision Transformer representations with topological descriptors for histopathological cancer classification. The proposed approach combines 768-dimensional ViT features with 1,200 Betti-0 and Betti-1 features extracted from RGB and HSV representations. Experiments on the ICIAR2018 and UT-Osteosarcoma datasets demonstrated that the fusion of deep and topological representations consistently outperforms the individual ViT and TDA configurations. The proposed framework achieved accuracies of 94.10\% and 96.54\% in the respective SOTA comparisons, while the ablation experiments further demonstrated the effectiveness of ViT--TDA fusion, achieving 94.10\% and 96.54\% accuracy, respectively. These results highlight the complementary nature of learned visual and explicit topological representations and demonstrate the potential of deep-topological feature fusion for robust histopathological cancer classification.

\section{Future Work}
\label{sec:future}

Future work will focus on evaluating ABD-TDA on larger and more diverse multi-center histopathological datasets to assess its generalizability. We will also investigate computationally efficient topological feature extraction, additional color-space representations, and more advanced fusion strategies. Finally, incorporating explainability and clinically oriented validation will be explored to improve the transparency and practical applicability of the proposed framework.

\section*{Declarations}

\textbf{Funding} \\
The author received no financial support for the research, authorship, or publication of this work.

\vspace{2mm}
\textbf{Conflict of interest/Competing interests} \\
The authors declare no conflict of interest.

\vspace{2mm}
\textbf{Ethics approval and consent to participate} \\
Not applicable. This study did not involve human participants or animals, and publicly available datasets were used.

 \vspace{2mm}
\textbf{Acknowledgement} \\
 The authors utilized an online platform to check and correct grammatical errors and to improve sentence readability.

\vspace{2mm}
\textbf{Consent for publication} \\
Not applicable.

\vspace{2mm}
\textbf{Data availability} \\
The datasets used in this study are publicly available.

\vspace{2mm}
\textbf{Materials availability} \\
Not applicable.

\vspace{2mm}
\textbf{Code availability} \\
The source code used in this study is publicly available at 

\vspace{2mm}
\textbf{Author contributions} \\
 Faisal Ahmed downloaded the data, conceptualized the study, developed the code, performed the data analysis, and wrote the manuscript. Faisal Ahmed reviewed and approved the final version of the manuscript.

\clearpage

\bibliographystyle{elsarticle-num-names}

\bibliography{refs}

\end{document}